\documentclass[paper=a4,11pt,DIV=11,parskip=half]{scrartcl}

\usepackage[utf8]{inputenc}
\usepackage{amsmath,amssymb,amsthm}
\usepackage{graphicx}
\usepackage[ruled,vlined,linesnumbered]{algorithm2e}
\usepackage[breakable]{tcolorbox}
\usepackage{float}
\usepackage{hyperref}
\usepackage[style=vancouver,backend=biber]{biblatex}
\usepackage[margin=2.5cm]{geometry}
\usepackage{titling}

\everydisplay{\abovedisplayskip=8pt \belowdisplayskip=8pt \abovedisplayshortskip=8pt \belowdisplayshortskip=8pt}

\hypersetup{
    colorlinks=true,
    urlcolor=blue,
    linkcolor=blue,
    citecolor=blue
}

\title{Deep learning-guided evolutionary optimization for protein design}

\author{
    Erik Hartman\textsuperscript{1,*} \and 
    Di Tang\textsuperscript{1} \and 
    Johan Malmström\textsuperscript{1}
}

\begin{document}

\begin{tcolorbox}[
    colback=gray!10,
    colframe=gray!30,
    arc=6pt,
    boxrule=0.5pt,
    left=8pt,
    right=8pt,
    top=8pt,
    bottom=8pt,
    width=\textwidth
]

\maketitle

\begin{center}
    \includegraphics[width=0.6\textwidth]{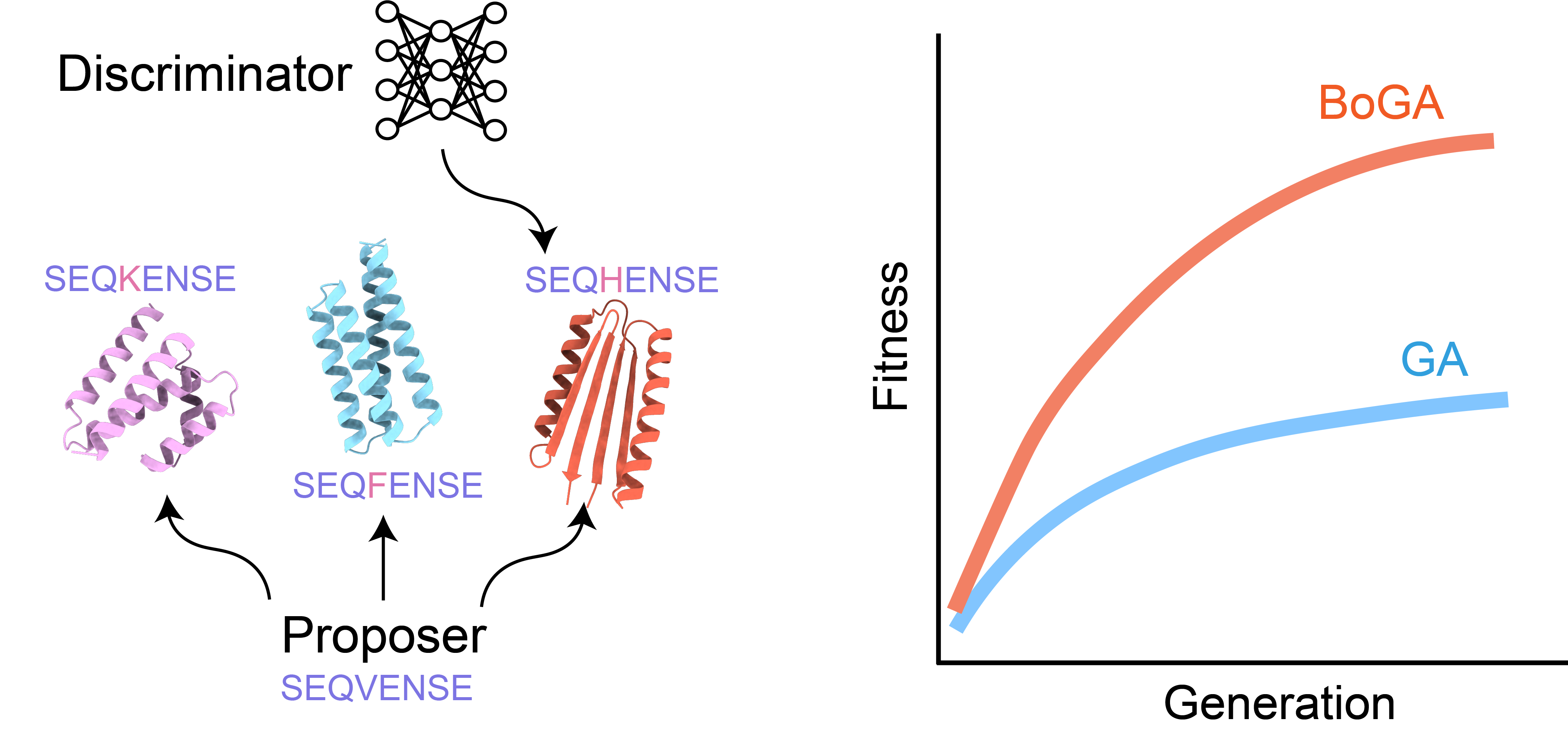}
\end{center}

\begin{abstract}
    \small
    Designing novel proteins with desired characteristics remains a significant challenge due to the large sequence space and the complexity of sequence-function relationships. Efficient exploration of this space to identify sequences that meet specific design criteria is crucial for advancing therapeutics and biotechnology. Here, we present BoGA (Bayesian Optimization Genetic Algorithm), a framework that combines evolutionary search with Bayesian optimization to efficiently navigate the sequence space. By integrating a genetic algorithm as a stochastic proposal generator within a surrogate modeling loop, BoGA prioritizes candidates based on prior evaluations and surrogate model predictions, enabling data-efficient optimization. We demonstrate the utility of BoGA through benchmarking on sequence and structure design tasks, followed by its application in designing peptide binders against pneumolysin, a key virulence factor of \textit{Streptococcus pneumoniae}. BoGA accelerates the discovery of high-confidence binders, demonstrating the potential for efficient protein design across diverse objectives. The algorithm is implemented within the BoPep suite and is available under an MIT license at \href{https://github.com/ErikHartman/bopep}{GitHub}.
\end{abstract}

\noindent\textsuperscript{1}Division of Infection Medicine, Faculty of Medicine, Lund University, Sweden

\noindent\textsuperscript{*}Corresponding author: Erik Hartman, \href{erik.hartman@med.lu.se}{erik.hartman@med.lu.se}

\end{tcolorbox}

\section{Introduction}
\normalsize

The design of proteins with specific characteristics, such as binding potential, structural properties, or catalytic activities holds transformative potential for next-generation therapeutics and biotechnology \cite{Fosgerau2015,Muttenthaler2021}. However, the astronomical size of sequence space combined with the complex, non-linear relationship between sequence and function makes rational design challenging. Identifying sequences that meet precise design criteria requires efficient navigation of this combinatorial landscape, where experimental validation of each candidate is both time-consuming and costly. To address this challenge, various computational methods have emerged to efficiently explore sequence space and identify candidates meeting specific design criteria \cite{Bhat2025,Leyva2025,Bennett2023,Jendrusch2025,bindcraft,boltzgen,Watson2023,fkhartman,Koh2025,Ebrahimi2023,Ingraham2023}.

Several methods have formulated protein design as a black-box optimization problem in sequence space and approached it using optimization algorithms. For example, genetic algorithms (GAs) use mutation, recombination, and selection to iteratively improve candidate sequences, mimicking natural evolution and have proven effective in a wide range of areas, including protein design \cite{BrowningJr2024,evobind,ChavesCarvalho2025,Schneider1998,Jendrusch2025,Hong2024,adalead}. While GAs can discover functional proteins, the efficiency is limited by the number of evaluations required to locate optimal solutions.

Another common optimization algorithm which has been applied to protein design is Bayesian Optimization (BO) \cite{Zong2023}. In BO, a surrogate model is trained on prior evaluations to approximate the objective function and guide the search towards promising regions of sequence space. When combined with GA, BO provides a principled way to prioritize candidates proposed by the GA based on prior evaluations and model predictions \cite{Hie2022,Liang2024,Wang2020,Hao2024,Freschlin2022}. While such algorithms, referred to as surrogate-assisted evolutionary algorithms or machine learning-guided directed evolution \cite{Tran2026}, have recently been proposed in the literature, they are yet to be applied to protein design, and their performance in this domain remains to be explored.

Here, we present \textbf{BoGA} (\textbf{B}ayesian \textbf{O}ptimization \textbf{G}enetic \textbf{A}lgorithm), a framework that combines evolutionary search with Bayesian optimization in an online learning loop. Building upon our previous development of \textbf{BoPep}, a Bayesian optimization framework for binder discovery \cite{Hartman2025}, BoGA extends this concept by incorporating a genetic algorithm as a stochastic proposal generator within a surrogate modeling loop. We demonstrate the utility of BoGA by first benchmarking its performance on simple sequence and structure design tasks. Then, we applied BoGA to design peptide binders against pneumolysin, a pore-forming toxin central to pneumococcal virulence \cite{Pereira2022}. BoGA is implemented within the BoPep suite, which provides modular support for embedding models, surrogate architectures, probabilistic inference modes, and docking-based evaluation functions. BoGA is available under an MIT license at \href{https://github.com/ErikHartman/bopep}{GitHub}.

\section{Main}

Classical genetic algorithms (GAs) for protein sequence design optimize a target property, such as antimicrobial potency or binding affinity, by iteratively mutating an evolving population of sequences. Those with the highest fitness over successive generations are selected, and the process repeats until convergence or a defined stopping criterion is reached. This incremental improvement is driven by stochastic mutation operators that introduce diversity and exploration into the sequence population. However, GAs typically require many evaluations of the objective function to identify high-performing sequences, which can be computationally expensive when evaluations involve structure prediction or docking simulations.

Bayesian optimization (BO) introduces a surrogate model trained on evaluated data to approximate the true objective function. This surrogate captures uncertainty in unobserved regions of sequence space and enables an acquisition function, such as expected improvement, to prioritize new candidates that balance exploitation (sampling near known high-performing sequences) and exploration (sampling uncertain regions).

Here, we combine these two paradigms in BoGA and apply them to the sequence design problem to find sequences that maximize or minimize a predefined objective (\textbf{Fig.~1}). In this formulation, the genetic algorithm serves as a proposal generator, producing diverse candidate sequences through stochastic mutation of the top-performing ones. The surrogate model is the discriminator, selecting which of the proposals that should undergo evaluation by defining the fitness of candidates through an acquisition function. As long as the evaluation of the objective function is more expensive than training the surrogate and generating proposals, and the surrogate model is reasonably accurate, this approach improves optimization efficiency. BoGA has been developed using state-of-the-art methods, including protein language models for sequence embedding, state-of-the-art probabilistic deep learning as surrogate models, and multiple structure prediction models to predict the 3-dimensional structure the sequence will fold to. 

A key feature of BoGA is the modularity, as multiple acquisition functions, surrogate architectures, embedding strategies, and selection strategies are supported. Below we outline a general setup for BoGA, followed by specific choices made in our experiments. We then demonstrate BoGA's performance on sequence- and structure-level optimization tasks, before applying it to the design of peptide binders against pneumolysin.

\subsection{Description of the BoGA framework}

To begin, let $\mathcal{X}$ denote the space of all possible peptide sequences of variable length $L$, where each sequence $\mathbf{x} \in \mathcal{A}^L$ with $\mathcal{A}$ representing the standard 20-letter amino acid alphabet. A design campaign begins from one or more seed sequences $\mathcal{X}_0 = \{\mathbf{x}_1, \dots, \mathbf{x}_{n_0}\} \subset \mathcal{X}$. At each generation $t$, BoGA maintains a dataset of evaluated sequences

\begin{equation}
\mathcal{D}_t = \{(\mathbf{x}_i, y_i)\}_{i=1}^{n_t}, \quad y_i = f(\mathbf{x}_i),
\end{equation}

where $f: \mathcal{A}^L \to \mathbb{R}$ is the objective function mapping a sequence $\mathbf{x}$ to a scalar score, such as binding affinity or docking energy in the case for binder design.

At each generation an elite subset $S_k$ is selected from either the current population $\mathcal{X}_t$ or the complete evaluation history $\mathcal{D}_t$ to serve as parents for mutation. A selection function $\mathcal{S}(\mathcal{X}_t, \mathcal{D}_t, k)$ determines $S_k$; for example, ranking candidates by $f(\mathbf{x})$ and selecting the top $k$ yields
\begin{equation}
S_k = \mathcal{S}(\mathcal{X}_t, \mathcal{D}_t, k) = \{\mathbf{x}_i : f(\mathbf{x}_i) \leq f(\mathbf{x}_j) \text{ for at most } k-1 \text{ other candidates}\},
\end{equation}
although alternative strategies such as t op-fraction selection, exponential sampling, or threshold-based filtering may be employed. The mutation operator $\mathcal{M}$ stochastically generates a diverse proposal pool of candidate sequences through mutation operators including deletion, insertion and substitution,
\begin{equation}
\mathcal{X}' = \{\mathbf{x}'_i \sim \mathcal{M}(S_k) : i = 1, \dots, k_{\text{propose}}\}.
\end{equation}
Each proposed sequence $\mathbf{x}' \in \mathcal{X}'$ is converted into a continuous embedding $\mathbf{z}' = \phi(\mathbf{x}')$ using a sequence encoder. These embeddings define the input space for the surrogate model $\hat{f}_{\boldsymbol{\theta}}$. The surrogate produces a predicted fitness $\hat{f}_{\boldsymbol{\theta}}(\mathbf{z}')$ for each embedded candidate.

To guide exploration, the surrogate outputs are transformed by an acquisition function $\alpha(\cdot)$, which prioritizes candidates expected to yield improvement either through high predicted performance or high model uncertainty. From the proposal pool, the $m_{\text{select}}$ candidates with the largest acquisition values are selected for explicit evaluation,
\begin{equation}
\mathbf{x}_{\text{next}} = \operatorname*{arg\,max}_{\mathbf{x}' \in \mathcal{X}'} \alpha\left(\hat{f}_{\boldsymbol{\theta}}(\phi(\mathbf{x}'))\right), \qquad |\mathbf{x}_{\text{next}}| = m_{\text{select}},
\end{equation}
and their true scores $f(\mathbf{x}_{\text{next}})$ are obtained from evaluation through e.g. structure prediction and computation of biophysical characteristics. These new evaluations expand the dataset used to train the surrogate, whose parameters are then updated by minimizing the predictive loss
\begin{equation}
\boldsymbol{\theta} \leftarrow  \operatorname*{arg\,min}_{\boldsymbol{\theta}} \mathcal{L}\left(\hat{f}_{\boldsymbol{\theta}}(\phi(\mathbf{x})), f(\mathbf{x})\right).
\end{equation}
The loop forms the core of the BoGA optimization cycle (Algorithm \ref{alg:boga}).

\begin{figure}[H]
\centering
\includegraphics[width=\textwidth]{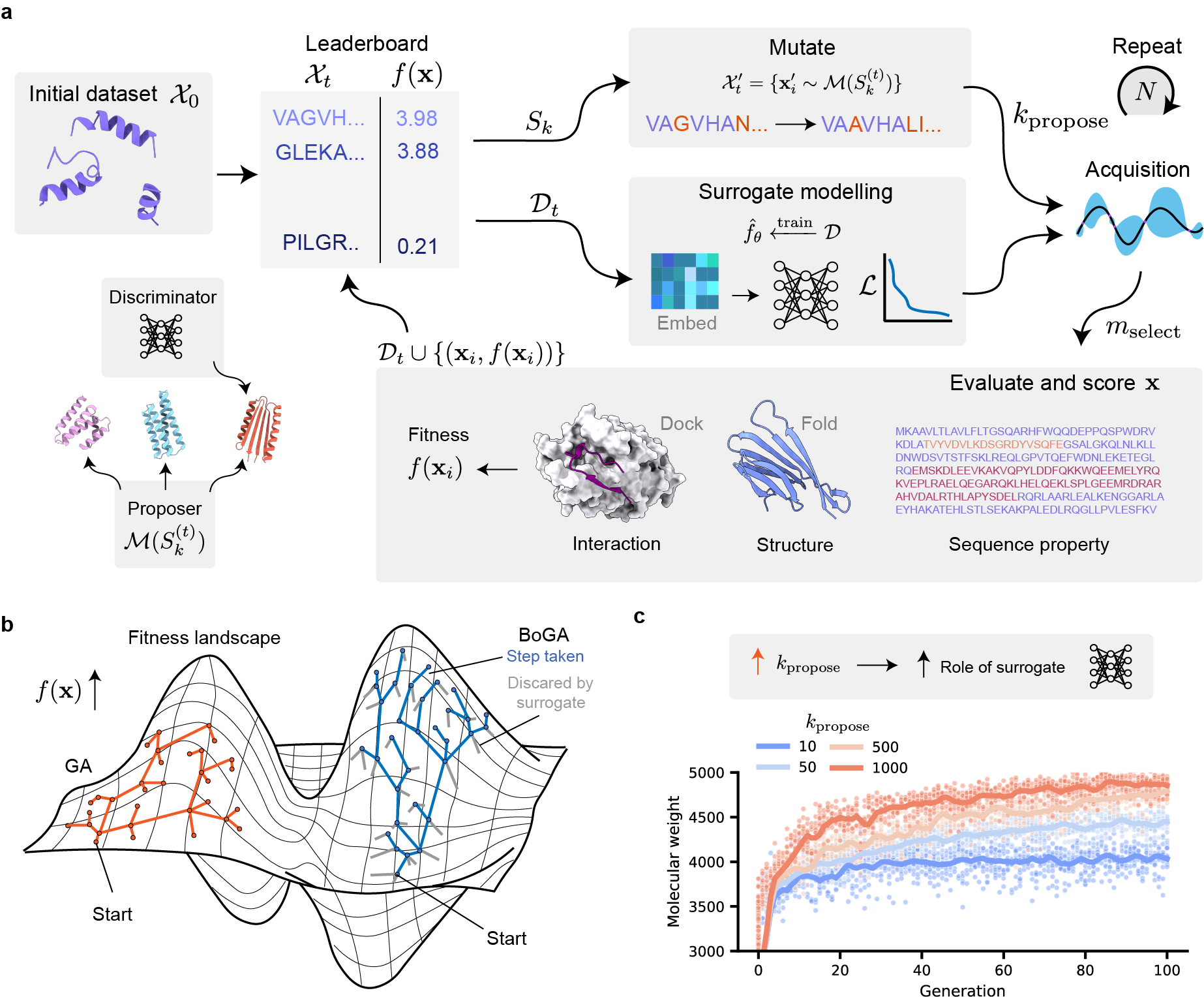}
\caption{\textbf{BoGA couples evolutionary optimization with Bayesian selection for sequence design.} \textbf{a} Schematic of the optimization cycle. From an initial evaluated dataset $\mathcal{D}_0$, an elite set $S_k$ is chosen from a leaderboard where each sequence is ranked by its fitness $f(\mathbf{x})$. A proposer (genetic mutations: substitutions/insertions/deletions) generates a pool of candidates $\mathcal{X}'$ of size $k_{\text{propose}}$. Sequences are embedded and scored by a surrogate model $\hat f_\theta$; an acquisition function $\alpha$ selects the $m_{\text{select}}$ top candidates for explicit evaluation. The fitness function is determined by the user, and can utilize properties derived from the amino acid sequence, the folded structure, and/or interchain interactions defined by a folded complex. The newly scored sequences are added to the dataset $\mathcal{D}_t$ and the loop is repeated. \textbf{b} Conceptual search in sequence space. A standard GA (orange) evaluates many proposals, including those that decrease fitness, whereas BoGA (blue) uses the surrogate to discard low-value proposals (gray) and concentrates evaluations to proposals that are more likely to increase fitness. \textbf{c} Example of optimization trajectories when maximizing molecular weight, showing improved performance as $k_{\text{propose}}$ increases.
}
\label{fig:boga}
\end{figure}

\begin{algorithm}[H]
\caption{BoGA: Bayesian Optimization Genetic Algorithm}
\label{alg:boga}
\KwIn{Initial sequences $\mathcal{X}_0$, objective function $f$, embedding model $\phi$, generations $T$, elite size $k$, proposal size $k_{\text{propose}}$, selection size $m_{\text{select}}$}
\KwOut{Optimized population $\mathcal{X}_T$}

Initialize dataset $\mathcal{D}_0 \leftarrow \{(\mathbf{x}, f(\mathbf{x})) : \mathbf{x} \in \mathcal{X}_0\}$\;
Initialize population $\mathcal{X}_0$\;

\For{$t = 1$ \KwTo $T$}{
    \tcp{Train surrogate model}
    $\boldsymbol{\theta}_t \leftarrow \arg\min_{\boldsymbol{\theta}} \mathcal{L}(\hat{f}_{\boldsymbol{\theta}}(\phi(\mathbf{x})), f(\mathbf{x}))$ for $(\mathbf{x}, f(\mathbf{x})) \in \mathcal{D}_{t-1}$\;
    
    \tcp{Select elite sequences (e.g., by ranking)}
    $S_k \leftarrow \mathcal{S}(\mathcal{X}_{t-1}, \mathcal{D}_{t-1}, k)$\;
    
    \tcp{Generate proposals via mutation}
    $\mathcal{X}' \leftarrow \emptyset$\;
    \For{$i = 1$ \KwTo $k_{\text{propose}}$}{
        $\mathbf{x}' \sim \mathcal{M}(S_k)$ \tcp*{Apply mutation operator}
        $\mathcal{X}' \leftarrow \mathcal{X}' \cup \{\mathbf{x}'\}$\;
    }
    
    \tcp{Score proposals with acquisition function}
    \For{$\mathbf{x}' \in \mathcal{X}'$}{
        $\mathbf{z}' \leftarrow \phi(\mathbf{x}')$ \tcp*{Embed sequence}
        $a(\mathbf{x}') \leftarrow \alpha(\hat{f}_{\boldsymbol{\theta}_t}(\mathbf{z}'))$ \tcp*{Compute acquisition value}
    }
    
    \tcp{Select top candidates for evaluation}
    $\mathcal{X}_{\text{eval}} \leftarrow \operatorname*{arg\,top}_{\mathbf{x}' \in \mathcal{X}'}^{m_{\text{select}}} a(\mathbf{x}')$\;
    
    \tcp{Evaluate selected candidates}
    \For{$\mathbf{x} \in \mathcal{X}_{\text{eval}}$}{
        $y \leftarrow f(\mathbf{x})$ \tcp*{Expensive evaluation}
        $\mathcal{D}_t \leftarrow \mathcal{D}_{t-1} \cup \{(\mathbf{x}, y)\}$\;
        $\mathcal{X}_t \leftarrow \mathcal{X}_{t-1} \cup \{\mathbf{x}\}$\;
    }
}

\Return{$\mathcal{X}_T$}\;
\end{algorithm}

\subsection{BoGA for sequence and structure optimization}

BoGA contains several parameters that can be tuned for different performances. One parameter is the number of proposals generated by the mutation function, $k_{\text{propose}}$ in contrast to the number of evaluated candidates, $m_{\text{select}}$. When these two are the same, the method is a standard genetic algorithm. However, as the $k_{\text{propose}}$ becomes larger than $m_{\text{select}}$, the surrogate model assumes a progressively larger role in discarding candidates. We first examined how $k_{\text{propose}}$ impacted optimization performance for simpler sequence specific properties such as $\beta$-sheet fraction and the normalized hydrophobic moment (uHrel). Here, the predicted $\beta$-sheet fraction at the amino acid level was defined as the fraction of E, M, A, and L residues in the sequence. ESM-2 was used to embed the sequences to a continuous latent space, PCA was used to reduce the dimensions of the embeddings to 100 dimensions, and a deep evidential regression BiGRU was used as the surrogate model. $m_{\text{select}}$ was fixed at $10$, while $k_\text{propose}$ was varied in $[10, 50, 100, 500, 1000]$. uHrel was a more complex property for the surrogate model to predict, resulting in validation $R^2 \approx 0.8$, while the beta sheet propensity was predictable with validation $R^2 \approx 1$. In both cases, the performance increased with $k_{\text{propose}}$, although sheet-fraction optimization benefitted more than the optimization for uHrel (\textbf{Fig.~2}). 

Optimization techniques that do not require pre-training offer flexibility in defining the optimization objective, enabling the optimization goal to be altered to guide generation towards any desired metric. To demonstrate this, we applied the algorithm to design proteins with specific secondary structures. AlphaFold 2 \cite{Evans2021} was used for structure prediction and the optimization goal was the secondary structure computed using DSSP multiplied by the predicted TM-score, optimizing for both structure and model confidence. We compared setting $k_\text{propose}$ to $10$, resulting in a standard GA, and to $500$, making the surrogate model select which $2\%$ of sequences to evaluate. In these cases, increasing $k_{\text{propose}}$ improved the optimization performance, increasing the mean fitness of the samples in the final 10 generations from $0.178 \pm 0.748$ to $0.253 \pm 0.0750$ for the sheet objective and from $0.367 \pm 0.578$ to $0.507 \pm 0.105$ for the helix objective (\textbf{Fig.~2c}). This demonstrates how BoGA can be used to optimize for arbitrary objectives and how the surrogate model can improve optimization efficiency by prioritizing candidates for evaluation.

\begin{figure}[H]
\centering
\includegraphics[width=\textwidth]{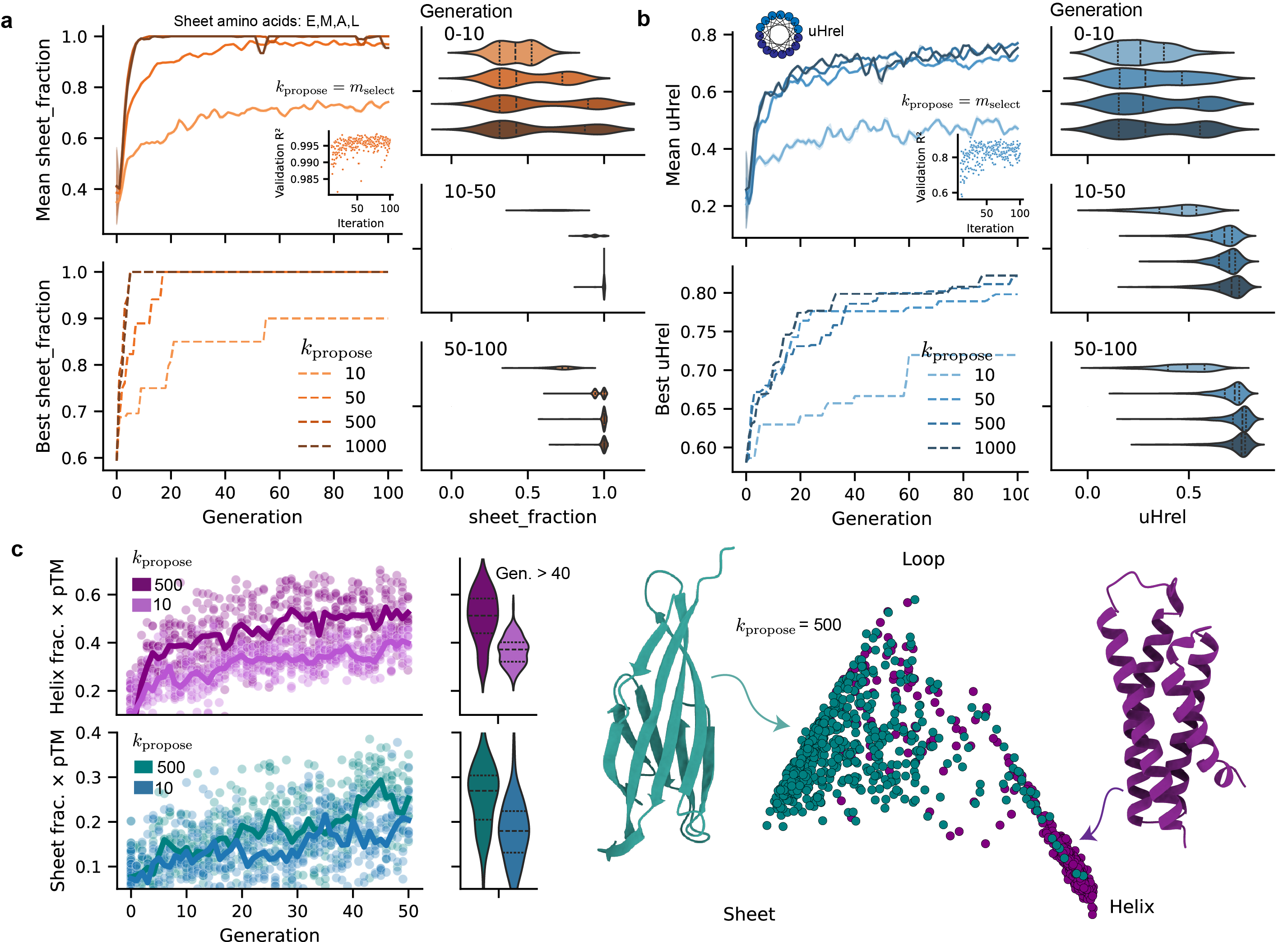}
\caption{\textbf{BoGA improves optimization efficiency for sequence- and structure-level objectives.}
\textbf{a} Optimization of $\beta$-sheet fraction (fraction of E, M, A, and L residues). The left panel shows optimization trajectories for different values of $k_{\text{propose}}$, and the right panel shows the distribution of fitness values at various stages of optimization. The inset shows the surrogate model $R^2$ over the course of optimization.
\textbf{b} Similar layout as in $\textbf{a}$ but the objective function is the normalized hydrophobic moment (uHrel).
\textbf{c} Structure-guided optimization using AlphaFold 2-predicted secondary structure weighted by predicted TM-score (pTM). The left panel shows optimization trajectories and violinplots show the distribution of fitness values for the last 10 generations. The right panel shows a triplot of the secondary structures of samples for the run with $k_{\text{propose}} = 500$.}
\label{fig:efficiency}
\end{figure}

\subsection{Generating binders against a bacterial virulence factor}

To evaluate whether BoGA can generate biologically relevant binders, we designed binders targeting pneumolysin (PLY), a pore-forming virulence factor central to \textit{Streptococcus pneumoniae} pathogenesis \cite{Pereira2022}. PLY mediates cytotoxicity by binding, via domain 4, to cholesterol-rich host cell membranes, and subsequently oligomerizing into large $\beta$-barrel pores, resulting in cell lysis. Neutralization of PLY therefore represents a promising therapeutic strategy with the potential to reduce both bacterial virulence and host-driven immunopathology. In previous work, we identified an antibody that potently neutralize PLY by binding to an epitope located in domain 4 \cite{Tang2026}. Here, we targeted the same epitope using peptide binders designed via BoGA (\textbf{Fig.~3a}).

The optimization objective was a previously developed scalar combination of structural prediction confidence metrics and biophysical measures of binding favorability \cite{Hartman2025}. The expected improvement acquisition function was used which balances explotation and exploration. Similarly to before, ESM-2 was used in combination with PCA to embed the peptide sequences, and a deep evidential regression \cite{der} BiGRU was used as a surrogate model. Hyperparameters were only optimized once after initialization, as prior experiments showed that optimizing them during later generations impaired performance, likely due to the homogeneity of the dataset. Boltz-2 was used to predict the conformation \cite{Passaro2025} of the peptide-protein complexes. The parameter $k_{\text{propose}}$ was set to 500, and $m_{\text{pool}}$ to 10. Optimization proceeded for 100 generations, and was initialized with 100 sequences. For comparison, we also performed optimization using $k_{\text{propose}} = 10$, to quantify the impact of the surrogate model on binder design. 

The run with $k_{\text{propose}} = 500$ accelerated discovery of high-scoring binders (\textbf{Fig.~3b}) and exhibited a more rapid increase in fitness across the population, indicating more efficient exploration of promising regions of sequence space. The resulting fitness distribution became bimodal as optimization proceeded, which likely demonstrates that the model selected both high confidence, high scoring binders, and low confidence binders that scored highly under expected improvement. Furthermore, the larger proposal pool yielded more high-confidence binders, measured by predicted interface pTM (ipTM) and peptide predicted aligned error (PAE) (\textbf{Fig.~3c}).

To further refine the top candidates from BoGA, we employed a post-optimization pipeline involving sequence recovery using ProteinMPNN \cite{Dauparas2022} and structural relaxation with FastRelax (\textbf{Fig.~3d}) \cite{Bennett2023,Dauparas2022,pyrosetta,relax}. The top 100 BoGA sequences were subjected to three rounds of sequence recovery and relaxation, generating 400 refined candidates. The candidates were then re-docked to the full-length PLY structure and rescored. The refined designs clustered in the high-confidence region of pLDDT versus ipTM space, many with favorable Boltz-2 interface distance scores (\textbf{Fig.~3e}). Following filtering for quality, solubility, and structural consistency, a final set of 41 high-confidence binders was obtained (\textbf{Fig.~3f}). Representative predicted complexes of top-performing binders in complex with PLY is shown in \textbf{Fig.~3f}. Independent structure prediction using AlphaFold 3 \cite{af3} further supported the binding pose of the top binders. Both modelling approaches indicated a consistent binding mode, high interface confidence and low binding free energy ($\Delta G$) (\textbf{Fig.~3g}).

This application highlights the utility of BoGA for the efficient design of peptide binders against a biologically relevant target, demonstrating how surrogate modeling can accelerate the identification of high-confidence candidates.

\begin{figure}[H]
\centering
\includegraphics[width=\textwidth]{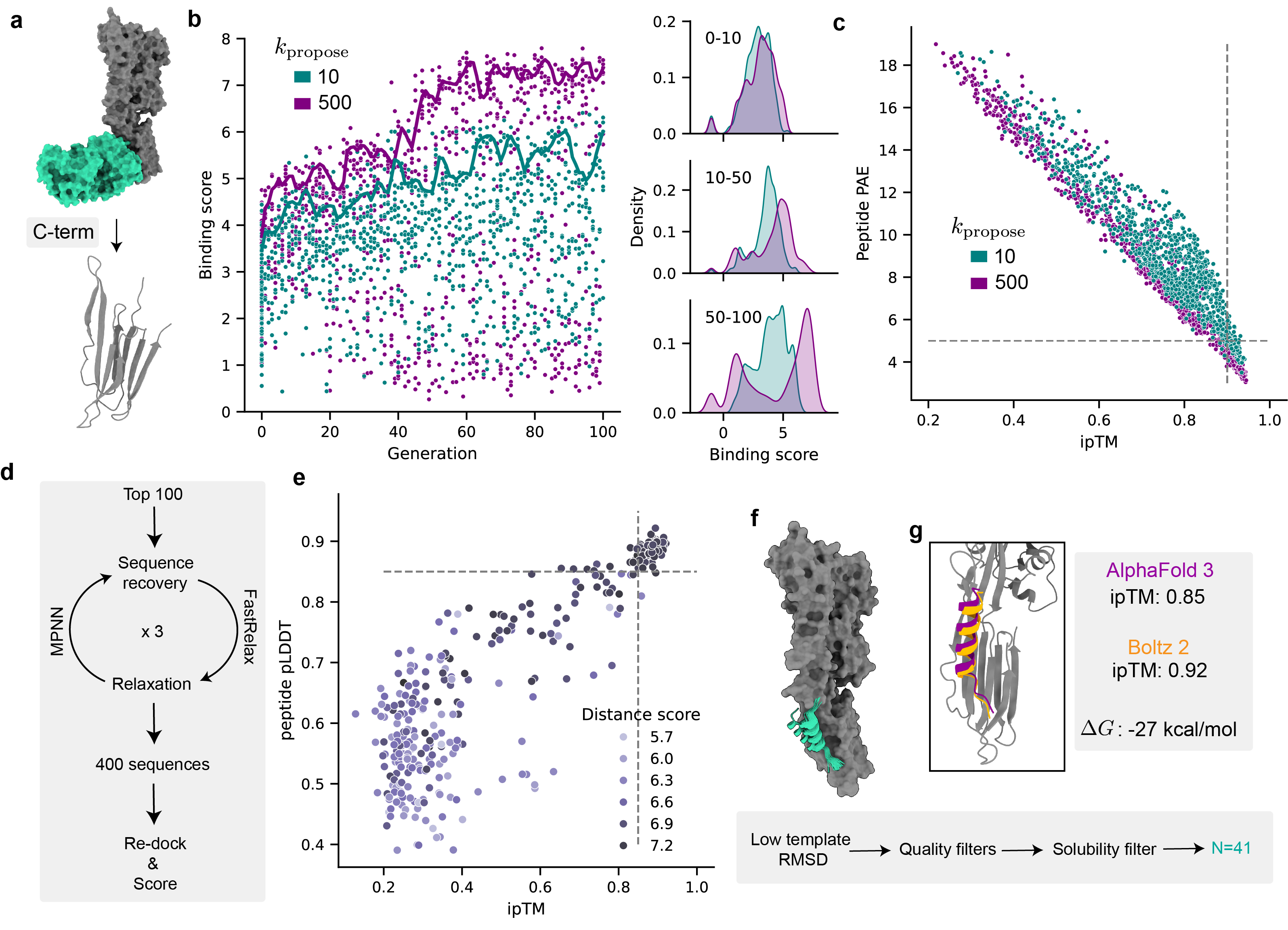}
\caption{\textbf{BoGA enables efficient design of peptide binders targeting pneumolysin.}
\textbf{a} Illustration of the neutralizing monoclonal antibody and PLY complex. PLY is shown in gray and the antibody in green. The model at the bottom represents domain 4.
\textbf{b} Optimization trajectories for the binder score across 100 generations for $k_{\mathrm{propose}} = 10$ (teal) and $k_{\mathrm{propose}} = 500$ (purple). The points indicate individual evaluated candidates; solid lines are running top quartile means. Right: Kernel density estimates of the binding score distribution during early (0-10), intermediate (10-50), and late (50-100) generations, showing that larger proposal pools accelerate discovery of high-scoring binders.
\textbf{c} Relationship between predicted interface pTM (ipTM) and peptide predicted aligned error (PAE) for evaluated sequences. Larger $k_{\mathrm{propose}}$ increases sampling of high-confidence, low-PAE binders. The dashed lines show cutoffs for ipTM=0.9 and PAE=5 Å.
\textbf{d} Outline of the post-optimization refinement approach. The top 100 BoGA sequences were subjected to three rounds of sequence recovery using ProteinMPNN and FastRelax, resulting in a total of 400 candidates, which are subsequently re-docked to the full-length PLY and scored.
\textbf{e} pLDDT versus ipTM for refined candidates. Points are colored by Boltz-2 interface distance score. High-confidence binders cluster at high ipTM and high pLDDT (upper right quadrant, dashed lines).
\textbf{f} Example predicted complexes (Boltz-2) of top-performing binders (cyan) bound to PLY (gray). Filtering of the top candidates using quality, solubility and structural consistency yielded a final set of 41 high-confidence binders.
\textbf{g} Orthogonal structure prediction using AlphaFold 3 and Boltz 2 for one of the top binders. Both models support a consistent binding pose with high interface confidence (ipTM = 0.85 for AlphaFold 3; ipTM = 0.92 for Boltz 2). Estimated binding free energy ($\Delta G$) from PyRosetta \cite{pyrosetta} indicates a strongly favorable interaction.}

\label{fig:pneumolysin}
\end{figure}

\section{Discussion}

BoGA integrates evolutionary optimization with Bayesian decision-making, resulting in an algorithm that efficiently explores sequence space. Unlike classical GAs, which evaluate all proposed candidates, BoGA uses a surrogate model to prioritize candidates for evaluation, allowing it to focus computational resources on the most promising sequences. Across sequence, structure, and binder-level objectives, this decoupling of \emph{generation} from \emph{evaluation} minimizes computational costs and enhances selection of high-scoring candidates.

Compared with other popular contemporary approaches such as hallucination-based approaches like BindCraft or diffusion-based approaches like RFDiffusion \cite{bindcraft,Watson2023}, BoGA offers several advantages. First, BoGA does not require training of a complex model on large datasets. Second, the optimization objective is flexible and can be adapted to any differentiable or non-differentiable metric without retraining. Third, BoGA can incorporate advances in structure prediction and docking in a flexible manner, as these models are only used for evaluation not generation. Lastly, BoGA is less constrained by the inductive biases of a generative model, as the genetic algorithm can explore sequence space more freely through mutation.

These advantages come at the cost of not having a pre-trained model capable of generating candidates in a single forward pass, which is computationally more efficient than iterative optimization. However, in typical pipelines structural prediction is performed to filter candidates, making evaluation, not generation, the primary bottleneck. In such contexts, the additional computational overhead of algorithms like BoGA is often justified.

The benefits of the surrogate model depend on the quality of the model. If the surrogate is poorly predictive or miscalibrated, acquisition scores become uninformative, resulting in behavior similar to a classical GA, or worse if the model introduces a systematic bias. Conversely, as surrogate accuracy and uncertainty calibration improve, BoGA becomes increasingly acquisition-driven and can approach near-deterministic optimization among the proposed candidates.

In general, BoGA is designed as a general and extensible strategy for protein design. The BoPep framework supports interchangeable sequence embeddings, surrogate architectures, acquisition functions, and mutation operators without altering the core optimization loop \cite{Hartman2025}. Overall, BoGA illustrates how coupling evolutionary proposals with Bayesian decision-making can enhance evolutionary algorithms for protein design. By leveraging surrogate models to prioritize candidates, BoGA accelerates discovery of high-performing sequences while maintaining flexibility across diverse design objectives.

\section{Methods}

Broadly, BoGA is a surrogate model-guided genetic algorithm for amino acid sequence design. Candidate sequences are iteratively generated by mutation, scored, and used to train a surrogate model that guides further exploration.

Let $\mathcal{X}$ denote the space of all possible peptide sequences of variable length $L$, where each sequence $\mathbf{x} \in \mathcal{A}^L$ with $\mathcal{A}$ representing the standard 20-letter amino acid alphabet. A design campaign begins from one or more seed sequences $\mathcal{X}_0 = \{\mathbf{x}_1, \dots, \mathbf{x}_{n_0}\} \subset \mathcal{X}$. At each generation $t$, BoGA maintains a dataset of evaluated sequences
\begin{equation}
\mathcal{D}_t = \{(\mathbf{x}_i, y_i)\}_{i=1}^{n_t}, \quad y_i = f(\mathbf{x}_i),
\end{equation}
where $f: \mathcal{A}^L \to \mathbb{R}$ is the objective function mapping a sequence $\mathbf{x}$ to a scalar score, such as binding affinity or docking energy in the case for binder design.

Each sequence $\mathbf{x}_i$ is embedded into a numerical representation
\begin{equation}
\mathbf{z}_i = \phi(\mathbf{x}_i),
\end{equation}
where $\phi: \mathcal{A}^L \to \mathbb{R}^D$ denotes an embedding model such as ESM-2 that maps sequences to a $D$-dimensional latent space. The embeddings $\mathbf{z}_i \in \mathbb{R}^D$ provide a continuous vector space for surrogate learning. To reduce computational complexity and mitigate overfitting when $D$ is large, dimensionality reduction techniques such as principal component analysis (PCA) or variational autoencoders (VAEs) may be applied to the embeddings prior to surrogate training.

A surrogate model $\hat{f}_{\boldsymbol{\theta}}$, parameterized by neural network weights $\boldsymbol{\theta}$, is trained to approximate $f$ by minimizing a loss between predicted and true scores:
\begin{equation}
\boldsymbol{\theta}_{t+1} = \arg\min_{\boldsymbol{\theta}} \mathcal{L}\big(\hat{f}_{\boldsymbol{\theta}}(\mathbf{z}_i), y_i\big), \quad (\mathbf{x}_i, y_i) \in \mathcal{D}_t.
\end{equation}
An elite subset is selected via a selection function $\mathcal{S}(\mathcal{X}_t, \mathcal{D}_t, k)$ that operates on either the current population $\mathcal{X}_t$ or the complete evaluation history $\mathcal{D}_t$:
\begin{equation}
S_k^{(t)} = \mathcal{S}(\mathcal{X}_t, \mathcal{D}_t, k).
\end{equation}
For instance, ranking sequences by fitness and selecting the top $k$ yields $S_k^{(t)} = \{\mathbf{x}_1, \dots, \mathbf{x}_k\}$ where $f(\mathbf{x}_1) < f(\mathbf{x}_2) < \cdots < f(\mathbf{x}_k)$. This elite subset is then selected for mutation. A stochastic mutation operator $\mathcal{M}$ perturbs $S_k^{(t)}$ to produce a proposal pool of size $k_{\text{propose}}$,
\begin{equation}
\mathcal{X}'_t = \{\mathbf{x}'_i \sim \mathcal{M}(S_k^{(t)}) : i = 1, \dots, k_{\text{propose}}\},
\end{equation}
where $\mathcal{M}$ introduces amino acid substitutions, insertions, or deletions with learned probabilities.

For each proposed sequence $\mathbf{x}' \in \mathcal{X}'_t$, the embedding $\mathbf{z}' = \phi(\mathbf{x}')$ is computed, and the surrogate predicts its fitness $\hat{f}_{\boldsymbol{\theta}}(\mathbf{z}')$. Candidates are then scored by an acquisition function $\alpha(\cdot)$, and the $m_{\text{select}}$ most promising candidates are chosen for evaluation:
\begin{equation}
\mathbf{x}_{\text{next}} = \operatorname*{arg\,max}_{\mathbf{x}' \in \mathcal{X}'_t} \alpha\big(\hat{f}_{\boldsymbol{\theta}}(\phi(\mathbf{x}'))\big).
\end{equation}
The true objective values are then obtained through physical evaluation, for example docking or co-folding, and the population and dataset are updated iteratively:
\begin{equation}
\mathcal{X}_{t+1} = \mathcal{X}_t \cup \mathbf{x}_{\text{next}}, \qquad \mathcal{D}_{t+1} = \mathcal{D}_t \cup \{(\mathbf{x}_i, f(\mathbf{x}_i)) : \mathbf{x}_i \in \mathbf{x}_{\text{next}}\}.
\end{equation}
This optimization constitutes the core of the BoGA algorithm. The ratio
\begin{equation}
\frac{m_{\text{select}}}{k_{\text{propose}}},
\end{equation}
governs the influence of the surrogate model: as the ratio decreases, the surrogate model assumes a greater role in guiding the search toward high-performing regions of sequence space.

BoGA is implemented within the BoPep framework \cite{Hartman2025}, which provides modular support for embedding models, surrogate architectures, probabilistic inference modes, and docking-based evaluation functions.

\subsection{Computational complexity}

The computational cost of BoGA comprises three main components: objective function evaluation, surrogate model training, and sequence embedding. For a standard genetic algorithm (GA) without surrogate modeling, the dominant cost per generation is the evaluation of $m_{\text{select}}$ candidates using the objective function $f$, yielding a per-generation complexity of $\mathcal{O}(m_{\text{select}} \cdot C_f)$, where $C_f$ denotes the cost of a single evaluation. Over $T$ generations, the total cost scales as $\mathcal{O}(T \cdot m_{\text{select}} \cdot C_f)$.

BoGA introduces additional overhead from surrogate modeling but aims to reduce the number of expensive evaluations required to reach convergence. At each generation, the surrogate model is trained on the growing dataset $\mathcal{D}_t$ at cost $C_{\text{train}}$, the $k_{\text{propose}}$ proposed candidates are embedded at cost $C_{\text{embed}}$, and the surrogate evaluates these embeddings at cost $C_{\text{surrogate}}$. The per-generation cost of BoGA is therefore $\mathcal{O}(C_{\text{train}} + C_{\text{embed}} + C_{\text{surrogate}} + m_{\text{select}} \cdot C_f)$.

The key advantage of BoGA emerges when $C_f \gg C_{\text{train}} + C_{\text{embed}} + C_{\text{surrogate}}$, that is, when the cost of evaluating the objective function far exceeds the combined cost of surrogate training, embedding, and inference. In such regimes, the surrogate can filter out low-quality candidates at relatively low computational cost, allowing the algorithm to focus expensive evaluations on promising regions of sequence space. This trade-off is particularly favorable in protein design applications where structure prediction is orders of magnitude more expensive than surrogate operations.

\subsection{\texorpdfstring{Evaluating the effect of $k_{\text{propose}}$}{Evaluating the effect of k-propose}}

To investigate the influence of proposal pool size on optimization efficiency, we varied $k_{\text{propose}}$ while fixing the number of expensive evaluations per generation ($m_{\text{select}}$). For each objective, runs were initialized with the same number of sequences and executed for a fixed number of generations. A GA baseline was implemented by setting $k_{\text{propose}} = m_{\text{select}}$ while keeping all other components identical.

\paragraph{Sequence optimization}
A simple sequence-level optimization was performed for two objectives: maximizing predicted $\beta$-sheet fraction (fraction of E, M, A, and L residues) and maximizing normalized hydrophobic moment (uHrel). ESM-2 was used to embed sequences, with PCA reducing the embeddings to 100 dimensions. A deep evidential regression BiGRU served as the surrogate model. For both objectives, $m_{\text{select}}$ was fixed at 10, while $k_{\text{propose}}$ was varied across [10, 50, 100, 500, 1000]. Each setting was run for 100 generations.

\paragraph{Structure optimization}
For structure-guided optimization, we applied the same $k_{\text{propose}}$ sweep at fixed $m_{\text{select}}$, but with objective evaluation requiring structure prediction. Candidate sequences were folded with AlphaFold 2 and secondary structure was assigned with DSSP. The structure objective was defined as the target secondary structure fraction multiplied by AlphaFold 2 pTM, thereby rewarding both structural content and prediction confidence.

\subsection{Application to pneumolysin}

Binder design was performed against pneumolysin (PLY) using BoGA with complex-structure-based evaluation. Optimization was initialized with 100 peptide sequences and run for 100 generations. The minimum peptide length was set to 8 and the maximum to 25. Sequences were embedded with ESM-2 and reduced to 100 dimensions by PCA. The surrogate model was a deep evidential regression BiGRU trained on all accumulated evaluations and used with an expected improvement acquisition function to select candidates from the proposal pool. The initial hyperparameter optimization was performed using 50 trials and 5-fold cross-validation, after which the model parameters were fixed for the remainder of the optimization.

For each selected peptide, the complex was predicted with Boltz-2, using 3 models and 5 recycles. Candidates were scored using a previously developed scalar objective that combines model confidence and interface favorability metrics (interface $\Delta G$, ipTM, a distance score, the peptide pAE, and the rosetta score) derived from the predicted complex. We compared a surrogate-dominant setting ($k_{\text{propose}}=500$, $m_{\text{select}}=10$) to a GA-like baseline ($k_{\text{propose}}=10$, $m_{\text{select}}=10$) to quantify the benefit of surrogate filtering in binder discovery. An example configuration used for the pneumolysin design campaign is provided in the Supplementary Information.

To refine top candidates, the best 100 BoGA sequences were subjected to three rounds of ProteinMPNN \cite{Dauparas2022} sequence recovery followed by Rosetta FastRelax \cite{pyrosetta,relax}, yielding 400 refined variants. Refined candidates were re-docked to the full-length PLY structure and rescored using the same evaluation pipeline, and final binders were selected by filtering for high interface confidence and overall structural quality. A subset of top candidates was assessed by complex prediction using AlphaFold 3.

\section{Acknowledgements}

We thank Jonas Wallin for helpful discussions surrounding notation and algorithm presentation. We also thank Malcolm Siljehag Alencar for helpful discussions and feedback on the algorithm. 

\section{Funding} 

EH was supported by The Royal Swedish Academy of Sciences (ME2024-0049). J.M. is part of the I-SPY network funded by Leducq Foundation for Cardiovascular Research. The project was supported by ERC (2024-ADG 101200871), Mats Paulssons foundation (2025-0042) and Alfred Österlunds Foundation. Views and opinions expressed are however those of the authors only and do not necessarily reflect those of the European Union or the European Research Council. Neither the European Union nor the granting authority can be held responsible for them.

\section{Code availability}
BoGA is available in the BoPep suite at \href{https://github.com/ErikHartman/bopep}{GitHub}.

\printbibliography[title=References]

\section{Supplementary Information}

An example of the BoGA configuration used for the pneumolysin binder design campaign is provided below:
\begin{verbatim}
    schedule = [
        {
            'acquisition': 'expected_improvement',
            'generations': 100,
            'm_select': 10,
            'k_pool': 5000,
            'acquisition_kwargs': {
                'top_fraction': 50,
                'selection_method': 'uniform',
            }
        },
    ]

    boga = BoGA(
        target_structure_path=target_structure_path,
        initial_sequences=starting_sequences,
        min_sequence_length=8,
        max_sequence_length=25,

        n_init=100,         
        mutation_rate=0.05,  
        
        surrogate_model_kwargs={
            'model_type': 'deep_evidential',
            'network_type': 'bigru',         
            'n_trials': 50,
            'n_splits': 5,                    
            'hpo_interval': 200, # will never happen
        },
        
        # Scoring configuration
        scoring_kwargs={
            'scores_to_include': [
                'interface_dG',            
                'iptm',          
                'in_binding_site',         
                'distance_score',        
                'peptide_pae',          
                'rosetta_score'         

            ],
            'binding_site_residue_indices': binding_site_residues,
            'required_n_contact_residues': 5,  
            'n_jobs': 12 
        },
        
        docker_kwargs={
            'models': ['boltz'],  
            'output_dir': ...
            'num_models': 3,            
            'num_recycles': 5,          
            'diffusion_samples': 3,      
            'recycling_steps': 5,
            'gpu_ids': ...
        },

        objective_function=bopep_objective_v1,
        objective_function_kwargs={},

        embed_method='esm',         
        embed_batch_size=64,        
        pca_n_components=100,      
        n_validate=0.2,
        
        log_dir = ...
        
    )
    boga.run(schedule=schedule)
\end{verbatim}

\end{document}